\documentclass[runningheads]{llncs}
\usepackage{graphicx}
\usepackage{float} 
\usepackage{siunitx}
\usepackage{longtable} 
\usepackage{booktabs} 
\usepackage{pgfplots}
\pgfplotsset{compat=1.6}
\usepackage{amsmath}
\DeclareMathOperator*{\argmax}{arg\,max}

\usepackage{comment}
\usepackage{slashbox}
\usepackage{caption}
\usepackage{subcaption}

\usepackage{hyperref}

\usepackage{footnote}
\makesavenoteenv{table}
\makesavenoteenv{tabular}

\pgfplotsset{soldot/.style={color=blue,only marks,mark=*}} \pgfplotsset{holdot/.style={color=blue,fill=white,only marks,mark=*}}
\begin{document}
\title{Improving accuracy and speeding up Document Image Classification through parallel systems}
\titlerunning{Improving accuracy and speeding up Document Image Classification}
%
%
\author{Javier Ferrando\inst{1} \and Juan Luis Domínguez\inst{1} \and Jordi Torres\inst{1,2} \and Raúl García\inst{1} \and \\David García\inst{1} \and Daniel Garrido\inst{3} \and Jordi Cortada\inst{4} \and Mateo Valero\inst{1,2}
}

%
\authorrunning{Ferrando et al.}
%
\institute{
  Barcelona Supercomputing Center - Centro Nacional de Supercomputación\\
  \email{\{javier.ferrando,juan.dominguez,jordi.torres,raul.garcia,\\david.garcia2,mateo.valero\}@bsc.es}
\and
Universitat Politècnica de Catalunya, UPC-BarcelonaTech\\
\and
Serimag Media - TAAD\\
\email{daniel\_garrido@serimagmedia.com}
\and
CaixaBank\\
\email{jorge.cortada@caixabank.com}
\vspace{-4mm}}
\maketitle              
\begin{abstract}
This paper presents a study showing the benefits of the EfficientNet models compared with heavier Convolutional Neural Networks (CNNs) in the Document Classification task, essential problem in the digitalization process of institutions. We show in the RVL-CDIP dataset that we can improve previous results with a much lighter model and present its transfer learning capabilities on a smaller in-domain dataset such as Tobacco3482. Moreover, we present an ensemble pipeline which is able to boost solely image input by combining image model predictions with the ones generated by BERT model on extracted text by OCR. We also show that the batch size can be effectively increased without hindering its accuracy so that the training process can be sped up by parallelizing throughout multiple GPUs, decreasing the computational time needed. Lastly, we expose the training performance differences between PyTorch and Tensorflow Deep Learning frameworks.


\end{abstract}

\section{Introduction} \label{sect:introduction}

Document digitization has become a common practice in a wide variety of industries that deal with vast amounts of archives. Document classification is a task to face when trying to automate their document processes, but high intra-class and low inter-class variability between documents have made this a challenging problem.

First attempts focused on structural similarity between documents \cite{layout_structural_similiarity} and on feature extraction \cite{tobacco3482,local_salient,doc_structure} to differentiate characteristics of each class. The combination of both approaches has also been tested \cite{clustering_separation}.

Several classic machine learning techniques have been applied to these problem, i. e. K-Nearest Neighbor approach \cite{nearest_neighbor}, Hidden Markov Model \cite{markov} and Random Forest Classifier \cite{unsupervised,tobacco3482} while using SURF local descriptors before the Convolutional Neural Networks (CNNs) came into scene.

With the rise of Deep Learning, researchers have tried deep neural networks to improve the accuracy of their classifiers. CNNs have been proposed in past works, initially in 2014 by Le Kang \textit{et al.} \cite{tobacco3482_v2} who started with a simple 4-layer CNN trained from scratch. Then, transfer learning was demonstrated to work effectively \cite{BigTobacco,deepdoc} by using a network pre-trained on ImageNet \cite{Imagenet}. And latest models have become increasingly heavier (greater number of parameters) \cite{CNNs_analysis,intra_domain,cutting_error} as shown in Table \ref{table:previous_architectures}, with the speed and computational resources drawback this entails.

Recently, textual information has been used by itself or as a combination together with visual features extracted by the previously mentioned models. Although Optical Character Recognition (OCR) is prone to errors, particularly when dealing with handwritten documents, the use of modern Natural Language Processing (NLP) techniques have demonstrated a boost in the classifiers performance \cite{noce,multimodal,stream}.

The contributions of this paper can be summarized in two main topics:
\begin{itemize}
    \item Algorithmic performance: we propose a model and a training procedure to deal with images and text that outperforms the state-of-the- art in several settings and is lighter than any previous neural network used to classify the BigTobacco dataset, the most popular benchmark for Document Image Classification (Table 1).
    
    \item Training process speed up: we demonstrate the ability of these models to maintain their performance while saving a large amount of time by parallelizing over several GPUs. We also show the performance differences between the two most popular Deep Learning frameworks (TensorFlow and Pytorch), when using their own libraries dedicated to this task.
\end{itemize}




\section{Document Image Classification}

Document Image Classification task tries to predict the class which a document belongs to by means of analyzing its image representation. This challenge can be tackled in two ways, as an image classification problem and as a text classification problem. The former tries to look for patterns in the pixels of the image to find elements such as shapes or textures that can be associated to a certain class. The latter tries to understand the language written in the document and relate this to the different classes.

\subsection{Datasets}


As mentioned earlier, in this work we make use of two publicly available datasets containing samples of images from scanned documents from USA Tobacco companies, published by Legacy Tobacco Industry Documents and created by the University of California San Francisco (UCSF). We find these datasets a good representation of what enterprises and institutions may face with, based on the quality and type of classes. Furthermore, they have been go-to datasets in this research field since 2014 with which we can compare results.

RVL-CDIP (Ryerson Vision Lab Complex Document Information Processing) is a 400.000 document sample (BigTobacco from now onwards) presented in \cite{BigTobacco} for document classification tasks. This dataset contains the first page of each of the documents, which are labeled in 16 different classes with equal number of elements per class.
A smaller sample containing 3482 images was proposed in \cite{tobacco3482} as Tobacco3482 (SmallTobacco henceforth). This dataset is formed by documents belonging to 10 classes not uniformly distributed.

\begin{small}
\begin{longtable}[]{@{}lrr@{}}
\captionsetup{skip=-11pt} 
\caption{Parameters of the CNNs architectures used in BigTobacco.}\\
\toprule
Model & \#Params\tabularnewline
\midrule
\endhead
AlexNet & 60.97M\tabularnewline
VGG-16 & 138.36M\tabularnewline
ResNet-50 & 25.56M\tabularnewline
Inception-V3 & 23.83M\tabularnewline
EfficientNet-B2 & \textbf{9.2M}\tabularnewline
EfficientNet-B0 & \textbf{5.3M}\tabularnewline
\bottomrule
\label{table:previous_architectures}
\end{longtable}
\end{small}
\vspace{-10.5mm}
\subsection{Deep Learning}

The proposed methods in this work are based on supervised Deep Learning, where each document is associated to a class (label) so that the algorithms are trained by minimizing the error between the predictions and the truth. Deep Learning is a branch of machine learning that deals with deep neural networks, where each of the layers is trained to extract higher level representations of the previous ones. These models are trained by solving iteratively an unconstrained optimization problem. In each iteration, a random batch of the training data is fed into the model to compute the loss function value. Then, the gradient of the loss function with respect to the weights of the network is computed (backpropagation) and an update of the weights in the negative direction of the gradient is done. These networks are trained until they converge into a loss function minimum.

\subsection{Computer Vision}
The field where machines try to get an understanding of visual data is known as Computer Vision (CV). One of the most well-known tasks in CV is image classification. In 2010 The ImageNet Large Scale Visual Recognition Challenge (ILSVRC) was introduced, a competition that dealt with a 1.2 million images dataset belonging to 1000 classes. In 2012 the first CNN-based model significantly reduced the error rate, setting the beginning of the explosion of deep neural networks. From then onwards, deeper networks have become the norm.

The most used architecture in Computer Vision have been CNN-based networks. Their main operation is the convolution one, which consists on a succession of dot products between the vector representations of both the input space ($L_{q} \times B_{q} \times d_{q}$) and the filters ($F_{q} \times F_{q} \times d_{q}$). We slide each filter around the input volume getting an \textit{activation map} of dimension $L_{q+1} = (L_{q} - F_{q} + 1)$ and $B_{q+1} = (B_{q} - F_{q} + 1)$. The output volume then has a dimension of $L_{q+1} \times B_{q+1} \times d_{q+1}$, where $d_{q+1}$ refers to the number of filters used. We refer to \cite{charu} (we used the same notation for simplicity) to a more detailed explanation. Usually, each convolution layer is associated to an activation layer, where an activation function is applied to the whole output volume. To reduce the number of parameters of the network, a pooling layer is typically located between convolution operations. The pooling layer takes a region $P_{q} \times P_{q}$ in each of the $d_{q}$ activation maps and performs an arithmetic operation. The most used pooling layer is the max-pool, which returns the maximum value of the aforementioned region.

\subsection{Natural Language Processing}

The features learned from the OCR output are achieved by means of Natural Language Processing techniques. NLP is the field that deals with the understanding of human language by computers, which captures underlying meanings and relationships between words.

The way machines deal with words is by means of a real values vector representation. Word2Vec \cite{word2vec} showed that a vector could represent semantic and syntactic relationships between words. CoVe \cite{mccann2017learned} introduced the concept of context-based embeddings, where the same word can have a different vector representation depending on the surrounding text. ELMo \cite{Peters:2018} followed Cove but with a different training approach, by predicting the next word in a text sequence (Language Modelling), which made it possible to train on large available text corpus. Depending on the task (such as text classification, named entity recognition...) the output of the model can be treated in different ways. Moreover, custom layers can be added to the features extracted by these NLP models. For instance,  ULM-Fit \cite{ulmfit} introduced a language model and a fine-tuning strategy to effectively adapt the model to various downstream tasks, which pushed transfer learning in the NLP field. Lately, the Transformer architecture \cite{NIPS2017_7181} has dominated the scene, being the bidirectional Transformer encoder (BERT) \cite{BERT} the one who established recently state-of-the-art results over several downstream tasks.

\section{Related Work}

Several ways of measuring models have been shown in the past years regarding document classification on the Legacy Tobacco Industry Documents \cite{tobacco_initial}. Some authors have tested their models on a large-scale sample BigTobacco. Others tried on a smaller version named SmallTobacco, which could be seen as a more realistic scale of annotated data that users might be able to find. Lastly, transfer learning from in-domain datasets has been tested by using BigTobacco to pre-train the models to finally fine-tune on SmallTobacco. Table \ref{Related_work} summarizes the results of previous works in the different categories over time.

First results in the Deep Learning era have been mainly based on CNNs using transfer learning techniques. Multiple networks were trained on specific sections of the documents \cite{BigTobacco} to learn region-based high dimensional features later compressed via Principal Component Analysis (PCA). The use of multiple Deep Learning models was also exploited by Arindam Das \textit{et al.} by using an ensemble as a meta-classifier \cite{intra_domain}. A VGG-16\cite{VGG16} stack of networks using 5 different classifiers has been proposed, one of them trained on the full document and the others specifically over the header, footer, left body and right body. The Multi Layer Perceptron (MLP) was the ensemble that performed the better. A committee of models but with a SVM as the ensemble was also proposed \cite{Roy}.

\begin{table}
\caption{Previous results comparison (accuracy in \%).}
\resizebox{\textwidth}{!}{%
\begin{tabular}{c|c|c|c|c|c|}
\cline{2-6}
& \textbf{BigTobacco} & \multicolumn{4}{c|}{\textbf{SmallTobacco}}                                              \\ \cline{2-6} 
&                     & \multicolumn{2}{c|}{BigTobacco Pre-training} & \multicolumn{2}{c|}{No Pre-training}     \\ \hline
\multicolumn{1}{|c|}{Author}                  & Image               & Image                  & Image + Text        & Image                & Image + Text      \\ \hline
\multicolumn{1}{|c|}{Kumar et al. (2014)\cite{tobacco3482}}     &                     &                        &                     & 43.8                 &                   \\ \hline
\multicolumn{1}{|c|}{Kang et al. (2014)\cite{tobacco3482_v2}}      &                     &                        &                     & 65.37                &                   \\ \hline
\multicolumn{1}{|c|}{Afzal et al. (2015)\cite{deepdoc}}     &                     &                        &                     & 77.6                 &                   \\ \hline
\multicolumn{1}{|c|}{Harley et al. (2015)\cite{BigTobacco}}    & 89.8                &                        &                     & 79.9                 &                   \\ \hline
\multicolumn{1}{|c|}{Csurka et al. (2016)\cite{Csurka}}    & 90.7                &                        &                     &                      &                   \\ \hline
\multicolumn{1}{|c|}{Noce et al. (2016)\cite{noce}}      &                     &                        &                     &                      & 79.8              \\ \hline
\multicolumn{1}{|c|}{Afzal et al. (2017)\cite{cutting_error}}     & 90.97               & 91.13                  &                     &                      &                   \\ \hline
\multicolumn{1}{|c|}{Tensmeyer et al. (2018)\cite{CNNs_analysis}} & 90.8               &                        &                     &                      &                   \\ \hline
\multicolumn{1}{|c|}{Das et al. (2018)\cite{intra_domain}}       & 92.21               &                        &                     &                      &                   \\ \hline
\multicolumn{1}{|c|}{Audebert et al. (2019)\cite{multimodal}}  &                     &                        &                     & 84.5                 & 87.8              \\ \hline
\multicolumn{1}{|c|}{Asim et al. (2019)\cite{stream}}      &                     & 93.2\footnote{Accuracy obtained in 9 classes that overlap in BigTobacco}                    & 95.8\footnote{Evaluation method not specified}             &                      &                   \\ \hline
\multicolumn{1}{|c|}{Proposed work (2020)}    & \textbf{92.31}      & \textbf{94.04}         &         94.9            & \textbf{85.99} & \textbf{89.47} \\ \hline
\end{tabular}}
\label{Related_work}
\end{table}

The addition of content-based information has been investigated on SmallTobacco by extracting text through OCR and embedding the obtained features into the original document images as a previous phase to the training process \cite{noce}. Lately, a MobilenetV2 architecture \cite{mobilenetv2} together with a CNN 2D \cite{cnn2d,wallace} taking as input FastText embeddings \cite{fast_text1,fast_text2} have achieved the best results in SmallTobacco \cite{multimodal}.

A study of several CNNs was carried out \cite{cutting_error}, where VGG-16 architecture was found optimal. Afzal \textit{et al.} also demonstrated that transfer learning from in-domain datataset like BigTobacco increases by a large margin the results in SmallTobacco. This was further investigated by adding content-based information with CNN 2D with ranking textual features (ACC2) to the OCR extracted.

As far as we are concerned, there is no study about the use of multiple GPUs in the training process for the task of Document Image Classification. However, parallelizing a computer vision task has been shown to work properly using ResNet-50, which is a widely used network that usually gives good results despite its low complexity architecture. Several training procedures are demonstrated to work effectively with this model \cite{15_min,1_hour}. A learning rate value proportional to the batch size, warmup learning rate behaviour, batch normalization, SGD to RMSProp optimizer transition are some of the techniques exposed in these works. A study of the distributed training methods using ResNet-50 architecture on a HPC cluster is shown in \cite{campos1,campos2}. To know more about the algorithms used in this field we refer to \cite{distributed}.

\section{Proposed Approach}
In this section we present the models used and a brief explanation of them. We also show the training procedure used in both BigTobacco and SmallTobacco and the pipeline of our approach to the problem.


\subsection{Image model}\label{sec:image_model}

EfficientNets \cite{efficientnet} are a set of light CNNs designed to scale up in a structured manner. The network's width (\textit{w}), depth (\textit{d}) and resolution (\textit{r}) are defined as: $w = \alpha^{\phi}$, $d = \beta^{\phi}$ and $r = \gamma^{\phi}$, where $\phi$ is the scaling compound coefficient. The optimization problem is set by constraining $\alpha \cdot \beta^{2} \cdot \gamma^{2} \approx 2$ and $\alpha \geq 1, \beta \geq 1, \gamma \geq 1$.

By means of a grid search of $\alpha$, $\beta$, $\gamma$ with AutoML MNAS framework \cite{mnasnet} and fixing $\phi=1$, a baseline model (B0) is generated optimizing FLOPs and accuracy. Then, the baseline network is scaled up uniformly fixing $\alpha$, $\beta$, $\gamma$ and increasing $\phi$. We find that scaling the resolution parameter as proposed in \cite{efficientnet} does not improve the accuracy obtained. In our experiments in Section \ref{results} we proceed with an input image size of $384\times384$, which corresponds to a resolution $r = 1.71$, as proposed by Tensmeyer \textit{et al.} in \cite{CNNs_analysis} with AlexNet architecture \cite{AlexNet}.

The main block of the EfficientNets is the mobile inverted bottleneck convolution \cite{mobilenetv2,mnasnet}. This block is formed by two linear bottlenecks connected through both a shortcut connection and an intermediate expansion layer with a depthwise separable convolution ($3\times3$) \cite{depthwise}.
Probabilities $P(class|FC)$ are obtained by applying the softmax function on top of the fully connected layer \textit{FC} of the EfficientNet model.

\subsubsection{Pre-training on BigTobacco}

We train EfficientNets (pre-trained previously on ImageNet) on BigTobacco using Stochastic Gradient Descent for 20 epochs with \textit{Learning Rate Warmup} strategy \cite{bag_tricks}, specifically we follow \textit{STLR} (Slanted Triangular Learning Rate) \cite{ulmfit} which linearly increases the learning rate at the beginning of the training process and linearly decreases it after a certain number of iterations. We chose the \textit{reference learning rate} $\eta$ following the formula proposed in \cite{1_hour} and used in \cite{15_min} and \cite{bag_tricks}. Specifically, we set $\eta = 0.2 \cdot \frac{nk}{256}$, where \textit{k} denotes the number of workers (GPUs) and \textit{n} the number of samples per worker. Figure \ref{fig:parallel_approach} shows the multi-GPU training procedure to get EfficientNet\textsubscript{BigTobacco}, which represents EfficientNet model pre-trained on BigTobacco. EfficientNet is loaded with ImageNet weights (EfficientNet\textsubscript{ImageNet}) and then located in different GPUs within the same node.

\subsubsection{Fine-tuning on SmallTobacco}\label{fine-tune}

\begin{figure}[H]
	\begin{centering}
	\includegraphics[width=1\textwidth]{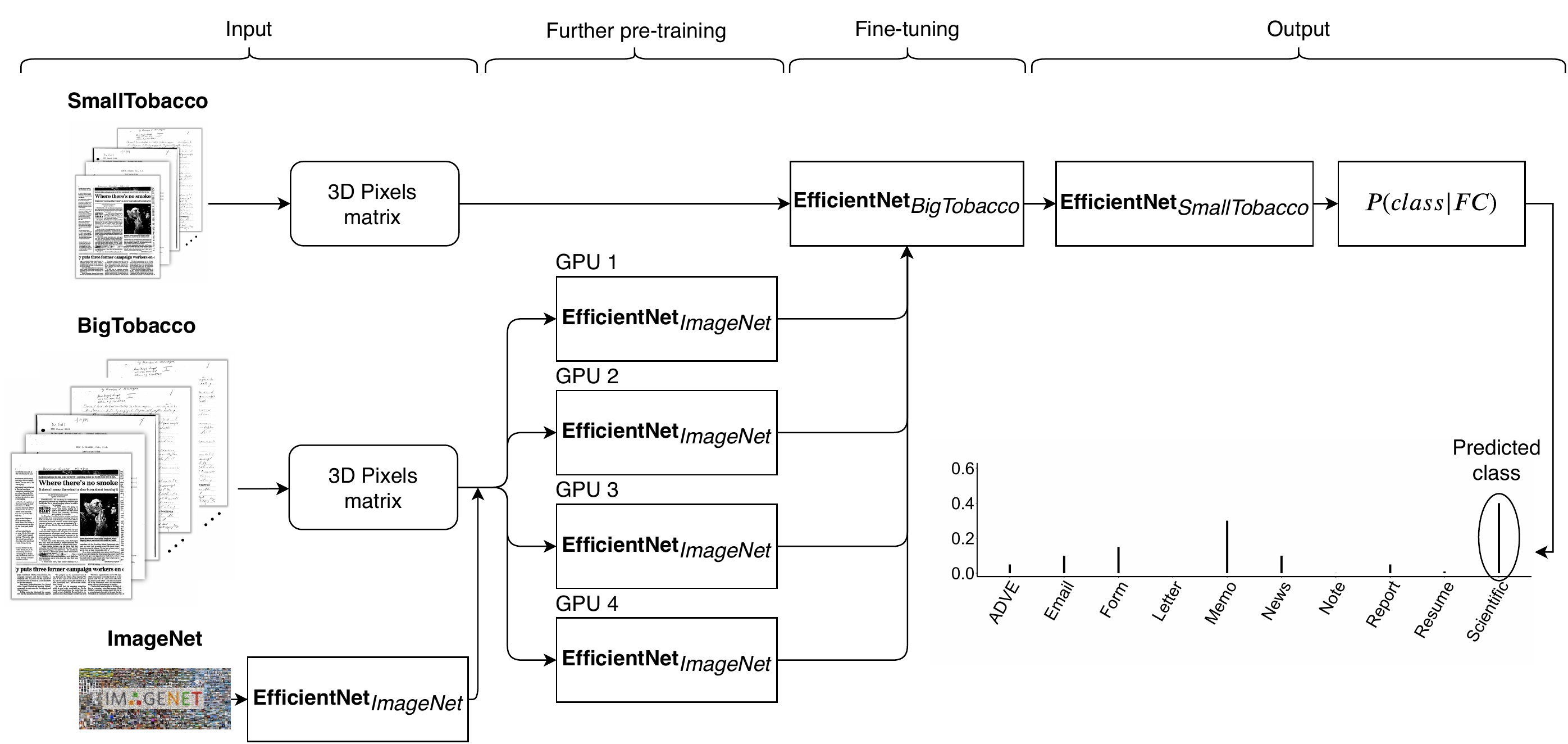}
	\caption{Pipeline of the different stages of the pre-training of EfficientNet over multiple GPUs.}
	\label{fig:parallel_approach}
	\end{centering}
\end{figure}

We fine-tune on SmallTobacco the pre-trained models by freezing the entire network but the last softmax layer. Just 5 epochs are enough to get the peak of accuracy. \textit{STLR} is used this time with $\eta = 0.8 \cdot \frac{nk}{256}$. Since only the last layer is trained, we reduce the risk of catastrophic forgetting \cite{catastrofic}. Final fine-tuned model is represented as EfficientNet\textsubscript{BigTobacco} in Figure \ref{fig:parallel_approach}.

\subsection{Text model}
Predictions from OCR Tesseract \cite{tesseract} are obtained by means of the BERT model \cite{BERT}. BERT is a multi-layer bidirectional Transformer encoder model pre-trained on a large corpus. In this work we use a modification of the original pre-trained BERT\textsubscript{BASE} version. In our case, we reduce to 6 the number of BERT layers since we find less variance in the final results and faster training/inference times. The output vector size is kept to 768. The maximum length of the input sequence is set to 512 tokens. The first token of the sequence is defined as $[CLS]$, while $[SEP]$ is the token used at the end of each sequence.

A fully connected layer is added to the final hidden state of the $[CLS]$ token $h_{[CLS]}$ of the BERT model, which is a representation of the whole sequence. Then, a softmax operation is performed giving $P(class|h_{[CLS]})$ the probabilities of the output vector $h_{[CLS]}$, i.e the whole input sequence, pertaining to a certain \textit{class}.

The training strategies used in this paper are similar to the ones proposed in \cite{tune_bert,classification_bert}. We use a learning rate $\eta_{B} = 3e^{-5}$ for the embedding, pooling and encoder layers while a a custom learning rate $\eta_{C}=1e^{-6}$ for the layers on top of the BERT model. A decay factor $\xi = 1e^{-8}$ is used to reduce gradually the learning rate along the layers, $\eta^{l}= \xi \cdot \eta^{l-1}$. ADAM optimizer with $\beta_1 = 0.9$ and $\beta_2 = 0.999$ and $L_{2}$-weight decay factor of 0.01 is used. The dropout probability is set at 0.2. Just 5 epochs are enough to find the peak of accuracy with a batch size of 6, the maximum we could use due to memory constraints.

\subsection{Image and Text ensemble}\label{ensemble}

\begin{figure}[H]
	\begin{centering}
	\includegraphics[width=1\textwidth]{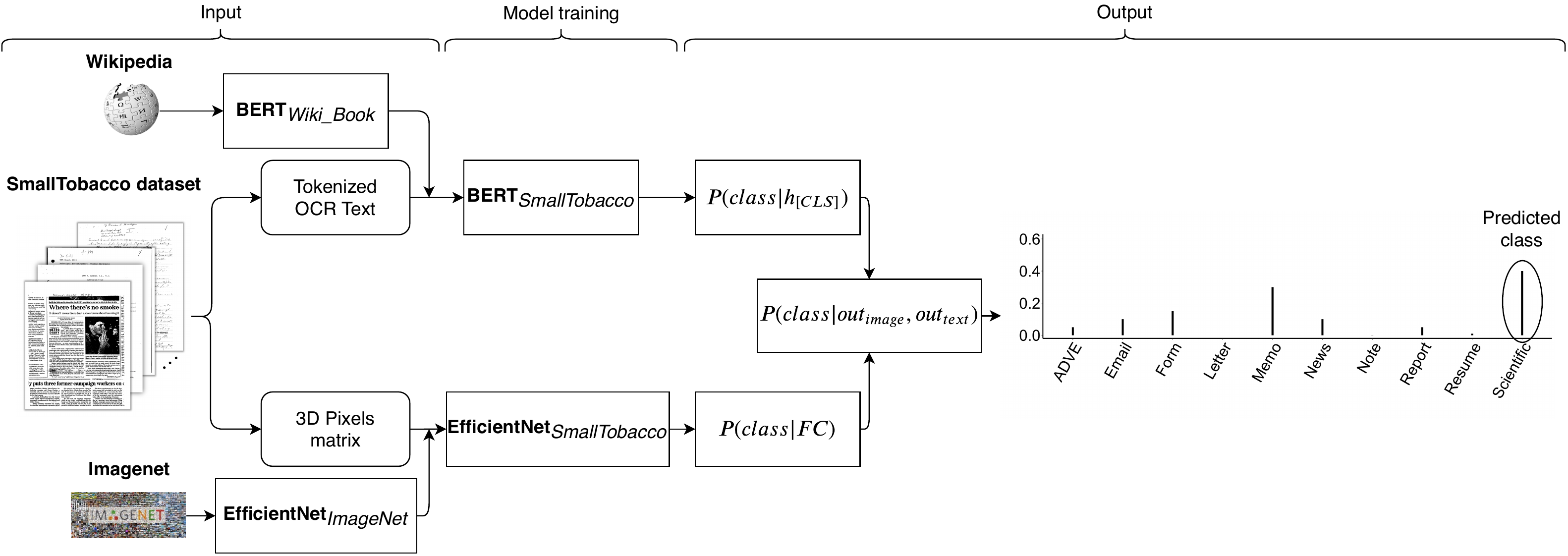}
	\caption{Pipeline of the proposed multimodal approach.}
	\label{fig:approach}
	\end{centering}
\end{figure}

In order to get the final enhanced prediction of the combination of both text and image model we use a simple ensemble as in \cite{stream}.
\[
 P(class|out_{image},out_{text}) = w_{1} \cdot P(class|h_{[CLS]}) + w_{2} \cdot P(class|FC)
\]
\[
Predicted\;Class = \argmax_{class}(P(class|out_{image},out_{text}))
\]
In this work $w_{1},w_{2} = 0.5$ are found optimal. These parameters could be found by a grid search where $\sum_{i=1}^{N}w_{i} = 1$, being $N$ the number of models. This procedure shows to be an effective solution when both models have a similar accuracy and it allows us to avoid another training phase \cite{multimodal}. In Figure \ref{fig:approach} this whole process is depicted.

\section{Results}\label{results}

In this section we compare the performance of the different EfficientNets in SmallTobacco and BigTobacco as showed in Table \ref{Related_work} and demostrate the benefits of the multiple GPU training. Experiments have been carried out using GPUs clusters Power-CTE\footnote{\urlstyle{same}\url{https://www.bsc.es/support/POWER_CTE-ug.pdf}} of the Barcelona Supercomputing Center - Centro Nacional de Supercomputación\footnote{\urlstyle{same}\url{https://www.bsc.es}}, each one composed by:
2 IBM Power9 8335-GTGH at 2.40GHz (20 cores and 4 threads/core), 512GB of main memory distributed in 16 dimms $\times$ 32GB at 2666MHz and 4 GPU NVIDIA V100 (Volta) with 16GB HBM2.

The operating system is RedHat Linux 7.4. The models and their training are implemented with PyTorch\footnote{\urlstyle{same}\url{https://pytorch.org/}} version 1.0 running on CUDA 10.1 and using cuDNN 7.6.4.

The only modification done to the images  is a resize to 384 $\times$ 384 as explained in Section \ref{sec:image_model} and, in order to avoid overfitting, a shear transformation of an angle $\theta \in [\ang{-5}, \ang{5}]$ \cite{CNNs_analysis} which is randomly applied in the training phase. No other modifications are used in our experiments. Source code is at \url{https://javiferran.github.io/document-classification}.

\subsection{Evaluation}
In order to compare with previous results in SmallTobacco dataset, we divide the dataset following the procedure in \cite{tobacco3482}. Documents are split in training, test and validation sets, containing 800, 2482 and 200 samples each one. 10 different splits of the dataset are created by randomly sampling from the 3482 documents, so that 100 samples per class are guaranteed between train and validation sets. In the Figure \ref{fig:finetuning} we give the accuracy on SmallTobacco as the median over the 10 dataset splits to compare with previous results. Accuracy on BigTobacco is shown as the one achieved on the test set. BigTobacco dataset used in Section \ref{sec:results_smalltobacco} is slightly modified, where overlapping documents with SmallTobacco are extracted. Top performing model's accuracies are written down in Table \ref{Related_work}.

\subsection{Results on BigTobacco}

\begin{figure}[H]
\centering
\begin{subfigure}{.6\textwidth}
  \centering
  \includegraphics[width=1\linewidth]{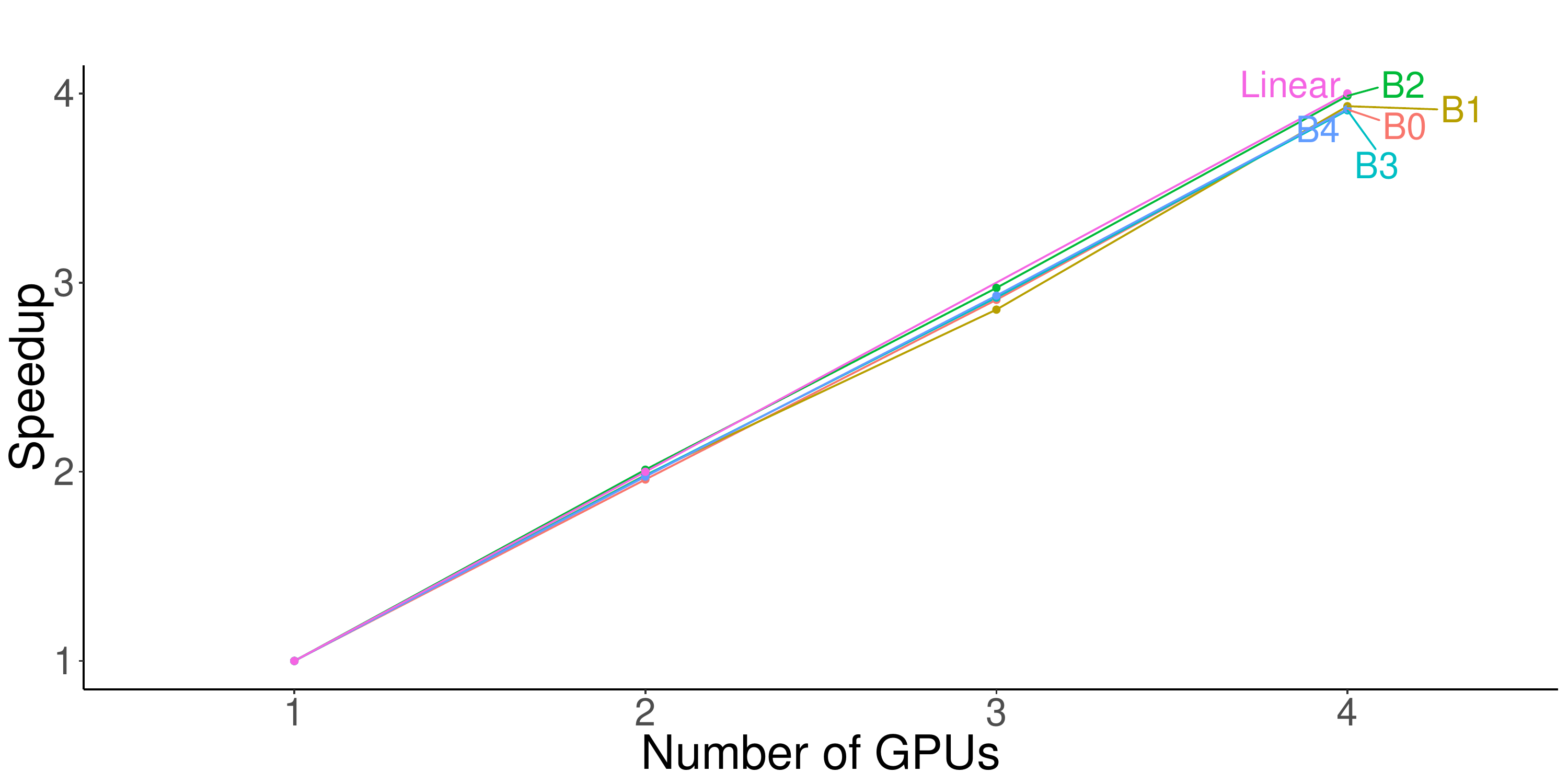}
\end{subfigure}%
\begin{subfigure}{.4\textwidth}
  \centering
  \includegraphics[width=0.9\linewidth]{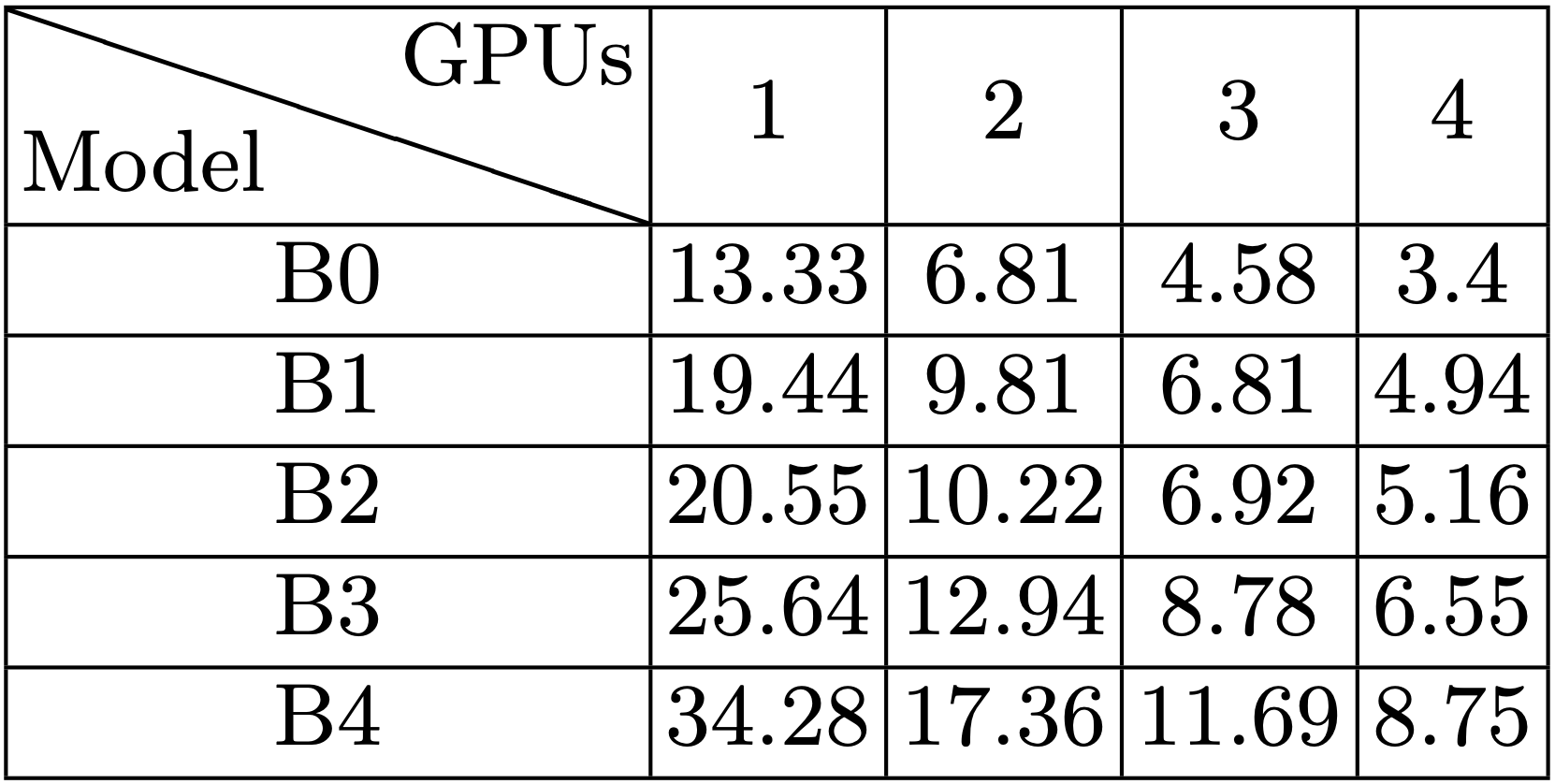}
\end{subfigure}
\caption{Left: speedup of the training process when parallelizing. Right: total time (hours) to train each model on different number of GPUs.}
\label{fig:times}
\end{figure}

We show in Figure \ref{fig:times} the time it takes to train the different networks while using 1, 2, 3 or 4 GPUs in a single node. In order to take advantage of the multiple GPUs we use data parallelism, which consists of placing a copy of the model in each of them. Since every GPU share parameters, it is equivalent to having a single GPU with a larger batch size.

The time reduction to complete the entire training process with B0 variant is $\approx 61.14\%$ lower when compared with B4 (4 GPUs). Time reduction by using multiple GPUs is clearly showed in the left plot of Figure \ref{fig:times}. For instance, EfficientNet-B0 benefits from a $\approx 75.4\%$ time reduction after parallelizing over 4 GPUs. The total training time of the EfficientNets on the different number of GPUs is showed in the right side of Figure \ref{fig:times}. The best performing model in BigTobacco dataset is EfficienNet-B4 with 92.31\% accuracy in the test set.

\subsection{Results on SmallTobacco}\label{sec:results_smalltobacco}

\begin{figure}
	\begin{centering}
	\includegraphics[width=1\textwidth]{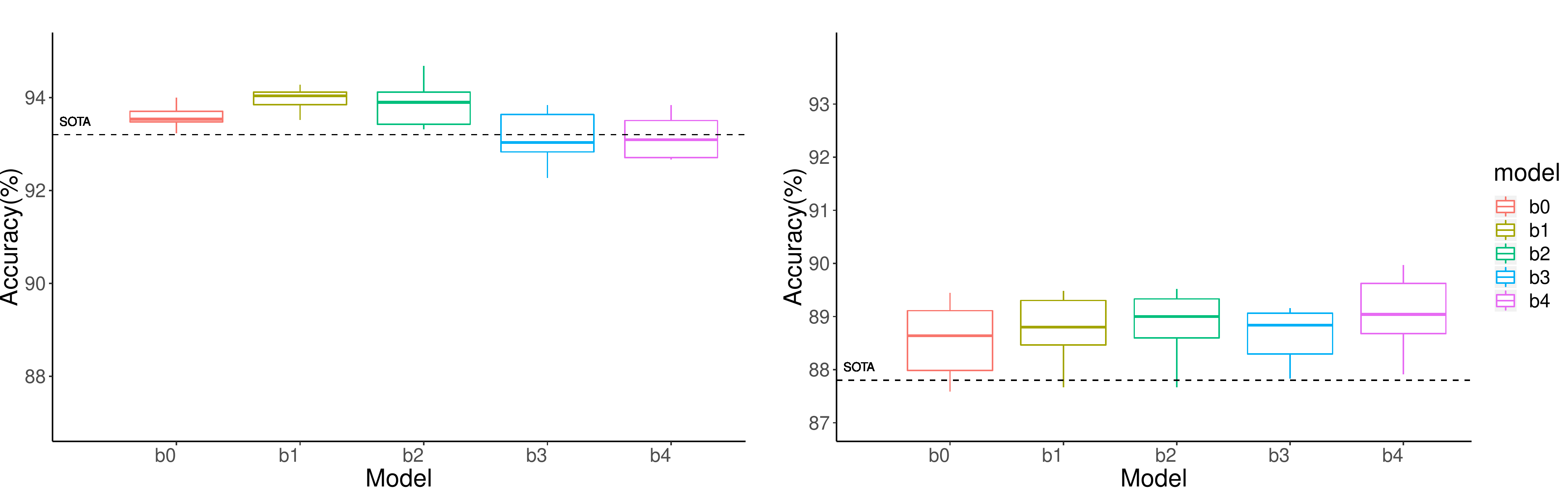}
	\caption{Accuracy obtained in SmallTobacco by models pre-trainined on BigTobacco (Left) and without BigTobacco pre-training (Right). Previous state-of-the-art (SOTA) results are shown with a horizontal dashed line.}
	\label{fig:finetuning}
	\end{centering}
\end{figure}

Accuracies of the EfficientNets pre-trained on BigTobacco and finally fine-tuned on SmallTobacco are depicted in the left plot of Figure \ref{fig:finetuning}. Simpler models perform with less variability between the 10 random splits than the heavier ones. The best performing model is the EfficientNet-B1, achieving a new state-of-art accuracy of 94.04\% median over 10 splits.

In this work, we also wanted to test the potential of light EfficientNet models on a small dataset such as SmallTobacco without the use of transfer learning from in-domain dataset, and compared it with the previous state-of-the-art. Results given by our proposed method described in Section \ref{ensemble} are shown in the right plot of Figure \ref{fig:finetuning}. Although we perform the tests over 10 different random splits to give a wider view of how these models work, in order to compare with \textit{Audebert et al.} \cite{multimodal} we calculate the average over 3 random splits, which gives us a 89.47\% accuracy.

Every ensemble model achieves better accuracy than previous results, and again, there is almost no difference between different EfficientNets results.

\subsection{Parallel platforms}

\begin{figure}
	\begin{centering}
	\includegraphics[width=1\textwidth]{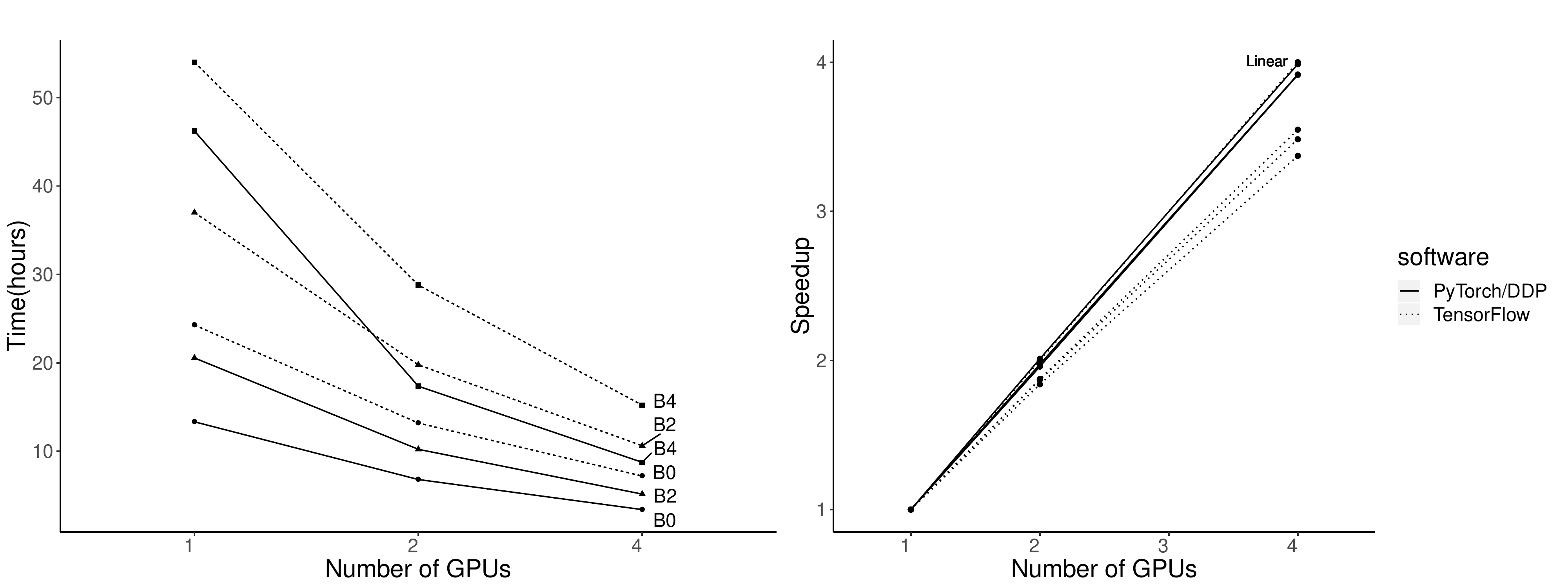}
	\caption{Left: the time to complete a whole training process. Right: speedup curves of TensorFlow and PyTorch.}
	\label{fig:tfvspy}
	\end{centering}
\end{figure}

Single GPU training requires a huge amount of time, especially when dealing with heavy architectures like in the case of the EfficientNet-B4, which takes almost two days to complete the whole training phase. For this reason, experimenting with several workers is crucial to minimize the amount of time spent on this tasks. We test the same model and training procedure with two of the main used frameworks to train Deep Learning models, PyTorch and Tensorflow\footnote{\urlstyle{same}\url{https://www.tensorflow.org/}}. In both cases we use their own APIs for making a synchronous distributed training in several GPUs by means of data parallelism, where training on each GPU is done in its own process. We use PyTorch's DistributedDataParallel and Tensorflow's tf.distribute.Strategy (tf.distribute.MirroredStrategy). In both libraries data is loaded from the disk to page-locked memory in each host, and from there to each GPU in a parallel fashion by means of multiple workers. Each GPU is ensured to get a minibatch with non overlapping data. Every GPU has an identical copy of the model and each one does its own forward pass. Finally, NCCL is utilized as a backend to run the all-reduce algorithm to compute the gradients in parallel between GPUs, before updating the model parameters. Since we have not been able to apply the shear transformation efficiently in Tensorflow, we show the results of both frameworks without that preprocess.
For this experiment we use the B0, B2 and B4 EfficientNets models.  The time it takes to train each model is showed on the left side of Figure \ref{fig:tfvspy}. PyTorch training is faster and the speedup more linear than in the case of TensorFlow. Some of this difference could be due to the data loading process, which we have not fully optimized in TensorFlow framework.

\section{Conclusion}
In this paper we have presented the use of EfficientNets for the Document Image Classification task and their scaling capabilities through several GPUs. By means of two versions of the Legacy Tobacco Industry Documents, a huge and a small dataset, we demonstrated the training process to obtain high accuracy in both of them. 
We have compared the different versions of the EfficientNets and raised the state-of-the-art classification accuracy to 92.31\% in BigTobacco and 94.04\% when fine-tuned in SmallTobacco. We can consider the B0 the best choice when considering limited computational resources.
We have also presented an ensemble method by adding the content extracted by OCR. A reduced version of the BERT model is trained and both models predictions are combined to achieve a new state-of-the-art accuracy of 89.47\%.

Finally, we have tested the same image models and training procedures in Tensorflow and PyTorch, where we have observed similar speedup values exploiting their libraries for distributed training. We have also tried distributed training in several GPU nodes by means Horovod framework \cite{sergeev2018horovod}, however the stack of software in our IBM Power 9 cluster is still in its early stages and we have not been able to obtain desired results. Nevertheless, future work may focus in testing this approach.

Future work may also evaluate the use of different OCR engines, as we suspect this could have a great impact on the quality of the text model predictions.

With this work we also want to provide to researchers a benchmark in the Document Image Classification task, which can serve as a reference point to effortlessly test parallel systems in both PyTorch and TensorFlow.

\section{Acknowledgements}
This work was partially supported by the Spanish Ministry of Science and Innovation and the European Regional Development Fund under contract TIN2015-65316-P, by the BSC-CNS Severo Ochoa program
SEV-2015-0493, and grant 2017-SGR-1414 by Generalitat de Catalunya and by the research agreement CaixaBank-BSC 2016-2021.

\normalsize


%
%
%
%

\bibliographystyle{splncs04}
\bibliography{main}

\end{document}